\documentclass[runningheads]{llncs}
\usepackage[pdftex]{graphicx}

\usepackage{tikz}
\usepackage{comment}
\usepackage{amsmath,amssymb} 
\usepackage{color}

\usepackage{booktabs}
\usepackage{tabularx}  
\usepackage{lipsum} 
\usepackage{mathtools}
\usepackage[linesnumbered,ruled,vlined]{algorithm2e}

\usepackage{enumitem}
\newlist{packed-enum}{itemize}{1}
\setlist[packed-enum]{leftmargin=*,label=\textbullet, itemsep=0.8pt, parsep=0.1pt}

\usepackage{makecell}

\makeatletter
\DeclareRobustCommand\onedot{\futurelet\@let@token\@onedot}
\def\@onedot{\ifx\@let@token.\else.\null\fi\xspace}

\def\eg{\emph{e.g}\onedot}

\def\etc{\emph{etc}\onedot}

\makeatother

\begin{document}
\newcommand\blfootnote[1]{%
  \begingroup
  \renewcommand\thefootnote{}\footnote{#1}%
  \addtocounter{footnote}{-1}%
  \endgroup
}
\newcommand{\mynew}[1]{{#1}}

\pagestyle{headings}
\mainmatter
\def\ECCVSubNumber{6946}  

\title{Telepresence Video Quality Assessment} 


%
\author{Zhenqiang Ying\inst{1} \and
Deepti Ghadiyaram\inst{2} \and
Alan Bovik\inst{1}}
%
%
\institute{\textsuperscript{$\ddagger$}University of Texas at Austin \and Facebook AI\\
\email{zqying@utexas.edu}, \email{deeptigp@fb.com},
\email{bovik@ece.utexas.edu}
}

%

\maketitle

\blfootnote{\textsuperscript{$\ddagger$} \mynew{The entity that conducted all of the data collection/experimentation.}
} 

\def\myenum{packed_enum}
\begin{abstract}

Video conferencing, which 
includes both video and audio content, has contributed to dramatic increases in Internet traffic, as the COVID-19 pandemic forced millions of people to work and learn from home.
 Global Internet traffic of video conferencing has dramatically increased 
Because of this, 
efficient and accurate video quality 
tools
are 
needed to monitor and perceptually optimize 
telepresence traffic streamed via Zoom, Webex, Meet, \etc.
However, existing 
models are limited in their prediction capabilities on multi-modal, live streaming telepresence content.
Here we address the significant challenges of Telepresence Video Quality Assessment (TVQA) in several ways. 
First, we mitigated the dearth of subjectively labeled data by collecting $\sim${}$2$k telepresence videos from different countries, on which we crowdsourced $\sim${}$80$k subjective quality labels.
Using this new resource, we created a first-of-a-kind online video quality prediction framework for live streaming,
using 
a multi-modal learning framework with separate pathways to compute visual and audio quality predictions.
Our all-in-one model is able to provide accurate quality predictions at the patch, frame, clip, and audiovisual levels. Our model achieves state-of-the-art performance on both existing quality databases and our new TVQA database, at a considerably lower computational expense, making it an attractive solution for mobile and embedded systems. 
\end{abstract}


\section{Introduction}

\begin{figure}[tbh]
\centering
\includegraphics[width=0.47\textwidth]{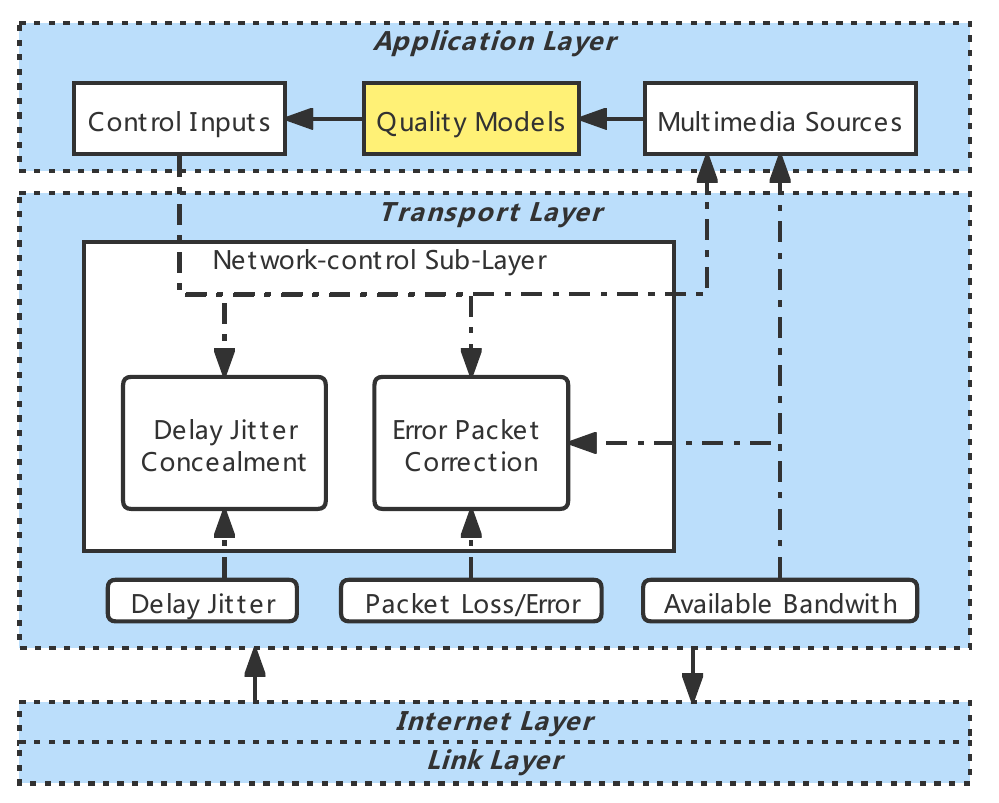}
\caption{
{\textbf{System-level view of quality models in the Internet protocol suite
(\textit{c.f.}~\cite{xu2017optimizing}).}}
Quality prediction algorithms provide feedback to control inputs which adjust buffering and rate control strategies at the network-control sub-layer, which 
determines how the multimedia data are packetized and transmitted, and how the received packets are buffered and corrected, with the goal of providing network conditions that satisfy the requirements of the application layer. }
\label{fig:system}
\end{figure}

Because of restrictions on physical meetings and travel necessary
to curb the spread of COVID-19, telecommuting and video conferencing are being utilized at unprecedented scales. 
For example, the revenues of Zoom, a popular videoconferencing platform, jumped by 355\% (year over year) in the $2^{nd}$ quarter of 2020,
after COVID-19 emerged~\cite{zoomStatistics}. 
Given the tremendous growth and continued vitality since then of telepresence videos 
for business, education, and interpersonal connections, 
being able to automatically monitor and control their perceptual quality has become quite important. 
Accurate and reliable 
quality predictors could be used to guide signal compression, transmission, reception, and display, providing perceptually optimized audio-visual conferencing experiences.
TVQA differs from other VQA/AQA problems, and presents particular challenges. 
To address it properly, it is necessary 
consider the overall telepresence workflow, 
including real-time requirements inherent in this type of error-prone, multi-directional live streaming. 
From a system-level view (\textit{c.f.} Fig.~\ref{fig:system}), video quality algorithms  serve a core purpose in the application layer, supporting 
tasks like acquisition, analysis, process evaluation, optimization of encoding, and stream monitoring during transmission and
reception~\cite{comprehensive_survey_avqa}. 
Achieving high perceptual quality of videoconferencing streams requires 
implementing 
multiple quality measurement tools
under limited bandwidth constraints, 
to obtain analytical data that can be used to tune  control inputs at run time ~\cite{tradeoffMultimediaApp}.
The main challenges of TVQA include: successfully integrating multimodal signal quality models, addressing the dearth of subjectively labeled data, and efficiently modeling live-streaming content quality.







We have addressed 
these challenges, by learning to efficiently model multimodel 
features and to provide various types of quality feedback in an online manner, using our first-of-a-kind telepresence video quality dataset. 
We summarize the contributions we make below:
\begin{packed-enum}
\item \textit{The first subjective database dedicated to telepresence  audio-visual quality}. 
We gathered $2320$ telepresence videos from YouTube and the Internet Archive.
By comparison, the number of unique contents is fewer than 100 in any prior audiovisual quality study.
We used Amazon Mechanical Turk to collect about $79K$ human perceptual quality judgments  on the collected content from more than $500$ subjects.


\item \textit{A first-of-a-kind online telepresence video quality prediction model, which we call Tele-VQA.} This new model can 
deliver rapid quality feedback as video streams arrive, by employing efficient backbones that extract multimodal features and integrate them into audio and visual regressors. 

\item {\textit{An all-in-one audio-visual quality framework that can process videos to deliver quality predictions at the patch, frame, clip, and audiovisual levels.} 
The image version of Tele-VQA, we call Tele-IQA, 
shows better performance 
than previous state-of-the-art models while requiring only 35\% parameters.
} 

\end{packed-enum}
\section{Background}\label{sec:background}



\noindent\textbf{A/VQA Models:} 
Video and audio quality assessment (VQA/AQA) algorithms can be classified into two main classes depending on whether reference pristine videos are available for comparison: full reference (FR) algorithms and no reference (NR) algorithms. 
A typical use case for NR models is the automated evaluation of user-generated content (UGC), such as videos uploaded to YouTube, Facebook, or TikTok. Unlike professional audio-visual content, UGC shared on prominent social media sites is typically acquired by novices having uncertain skills, 
using handheld cameras, with little or no editing. Similarly, telepresence videos are often captured by low-quality devices
under imperfect conditions, and are then subjected to 
compression, processing, and transmission artifacts before arriving on viewers' displays. 
Generally, no reference signal is available, so for the TVQA problem (both visual and audio signal evaluation), we only consider NR algorithms, since pristine contents are generally unavailable.

While there has been substantial progress on the development of top-performing learned models for NR-VQA (shallow models include~\cite{nrvqa1,nrvqa3,nrvqa5,nrvqa8,nrvqa10,nrvqa12,nrvqa13,tlvqm,videval,nrvqa-nstss,tu2021rapique}, and deep ones include ~\cite{vsfa,deepvqa,vmeon,zhangvqa,mlspvqa}) and NR-AQA~\cite{zheng2021towards_aqa,simou2020towards_aqa,warzybok2014subjective_aqa,gamper2019intrusive_aqa,goetze2010speech_aqa},
but relatively few NR-AVQA models~\cite{martinez2014no,annavqa,navidad,UnB2013} exist. 
Most existing AVQA models use ``handcrafted" statistical features to drive shallow learners (SVMs~\cite{navidad}, random forest ensembles~\cite{INRS}, \etc).
ANNAVQA~\cite{annavqa} was the first deep model proposed. It utilizes a pretrained convolutional
neural network (CNN) model to extract A/V features from single video frames and aligned short audio segments. Thus far, all existing audio-visual QA models, whether handcrafted or deep, 
operate only on video frames without computing spatiotemporal video features. 
These models focus on modular designs, without any system-level analysis of how to apply them to live streaming content, how to handle a missing modality, or how to supply multiple levels or abstractions of audio-visual quality. 

\noindent\textbf{A/VQA Datasets:} 
Representative databases are essential to learn effective VQA algorithms. 
A frequent misconception is that real images/videos can be characterized by one or two well-defined distortions.
In reality, there are several types and severities of distortions that often coexist, interact, and create new distortions~\cite{clive,paq2piq,livevqc,patchvq}.
Early VQA databases generally comprise a small number of unique source pristine videos (typically $10$-$15$), manually distorted by one of a few synthetic impairments (\eg, Gaussian blur, compression, or transmission artifacts)~\cite{epflPoli,livevqd,tum1080p,mcljcv,videoset,vqeg,csiq} (Table 1).  
These datasets are not representative enough to capture the complex characteristics of real-world telepresence videos.
More recent VQA databases have increased content diversity 
affected by authentic distortions~\cite{cvd2014,livequalcomm,konvid1k,livevqc,ytugc,patchvq,mlspvqa},
but none include audio signals.
Although several audiovisual QA databases have been released~\cite{hands2004basic,winkler2006perceived,VQEG_MM,UnB2013,martinez2018combining,martinez2014no}, 
the number of unique contents and degradation types they contain is quite limited, and do not reflect  real-world scenarios such as telepresence~\cite{UnB2013,UnB2018,you2010perceptual}. 
\begin{table*}[t!]

\caption{{Summary of Popular Audiovisual QA Databases}. VC: Video compression; TE: Transmission Errors; AC: Audio compression;
}
\scriptsize
\centering
\begin{tabular}{@{}clllccrr@{}}
\toprule
Year & Name         & \makecell{\# Unique \\ contents} & \makecell{\# Total \\ videos} & Distortion Type  & \# Annotators & \# Total Ratings \\ \midrule
2010 & PLYM~\cite{PLYM}          & 6                  & 60              & VC, TE            & 16            & 960              \\
2012 & VQEG-MM~\cite{VQEG_MM}               & 10                 & 60              & VC, AC                            & 35            & 2100             \\
2012 & TUM~\cite{tum1080p}                   & 5                  & 20              & VC, TE  & 21            & 420              \\
2013 & UnB-AVQ 2013~\cite{UnB2013}          & 8                  & 72              & VC, AC                 & 16            & 1, 152           \\
2016 & INRS~\cite{INRS,INRS2}                  & 1                  & 160             & VC, TE          & 30            & 4, 800           \\
2016 & MMSPG~\cite{MMSPG}                 & 9                  & 27              & Display devices                & 20            & 540              \\
2018 & UnB-AVQ Exp1~\cite{UnB2018}          & 60                 & 720             & VC, TE     & 60            & 43, 200          \\
2018 & UnB-AVQ Exp2~\cite{UnB2018}          & 40                 & 800             & VC, TE    & 40            & 32, 000          \\
2018 & UnB-AVQ Exp3~\cite{UnB2018}          & 40                 & 800             & VC, TE    & 40            & 32, 000          \\
2018 & LIVE-NFLX-II~\cite{LIVE_NFLIX2}          & 15                 & 420             & TE  & 65            & 27, 300          \\
2020 & LIVE-SJTU~\cite{min2020study}             & 14                 & 336             & VC, AC                   & 35            & 11, 760          \\
\textbf{2021} & \textbf{Proposed database}        & \textbf{2320}               & \textbf{2320}            & \textbf{In-the-wild}                 & \textbf{526}           & \textbf{78, 880} \\ \bottomrule
\end{tabular}
\vspace{-1em}
\label{tab:databases_summary}
\end{table*}

\section{TVQA Dataset and Human Study}

We collected $78, 880$ ratings ($34$ ratings on each video) 
on 2320 videos from $526$ subjects. 
As shown in Table~\ref{tab:databases_summary}, our telepresence quality  dataset is substantially larger than any previous subjective audiovisual dataset.  
Here we describe the new telepresence video quality dataset we constructed and the subjective quality study we conducted on it. 






\subsection{Data Acquisition}


\noindent\textbf{Data Sources:} ``Speaker-view'' and ``screen content'' are the two main types of contents encountered in telepresence videos.
The former can be found in widely used face analysis databases~\cite{chung2018voxceleb2,chung2016lip,ephrat2018looking}, but most of these are of television broadcasters, and 
are not affected by ``in-the-wild'' distortions typical of telepresence videos. 
Screen content video quality databases are available~\cite{cheng2020screen,li2020subjective} but audio signals are excluded in those studies. 
Therefore, we decided to crawl videos online instead of using videos from existing databases.
From among 6 million videos from the Internet
Archive, 
we filtered by relevant keywords and found about 7k recorded virtual meetings and 
randomly sampled 1129 videos from these to avoid content redundancies.
To further increase the distortion diversity, we manually searched YouTube videos uploaded from around 80 countries, using a location-based YouTube search engine~\cite{geofind}. In the end, we obtained 2,320 videos. 

\begin{figure}[t!]
\begin{center}
\includegraphics[ width=0.7\linewidth]{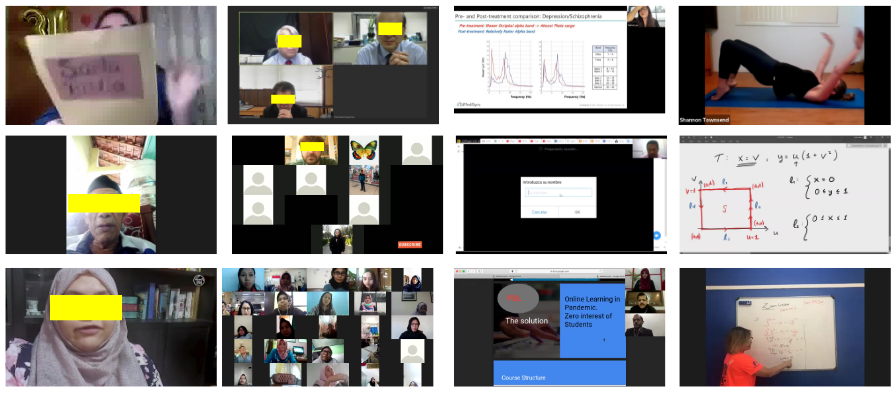}
\caption{{\textbf{Sample video frames from the proposed database}, each resized to fit. The actual videos are of highly diverse sizes and resolutions. Faces are masked to ensure privacy.}}
\label{fig:exemplar}
\end{center}
\end{figure}

\noindent\textbf{Data Processing:} Each video was randomly cropped to an average duration of 7 seconds using ffmpeg~\cite{ffmpeg}.
To keep the quality intact, 
we did not apply re-encoding, scaling, or any further processing that could affect perceptual quality. 
Fig.~\ref{fig:exemplar} shows 12 randomly selected video frames from the database.
It is evident that we obtained a highly diverse TVQA dataset, representative of telepresence content (including grid-views of multiple speakers, single speaker views, slide-sharing, screen content, etc.), resolution, aspect ratios, and distortions.


\subsection{Study Interface Design}



\begin{figure}[t]
\centering
\includegraphics[width=0.28\textwidth]{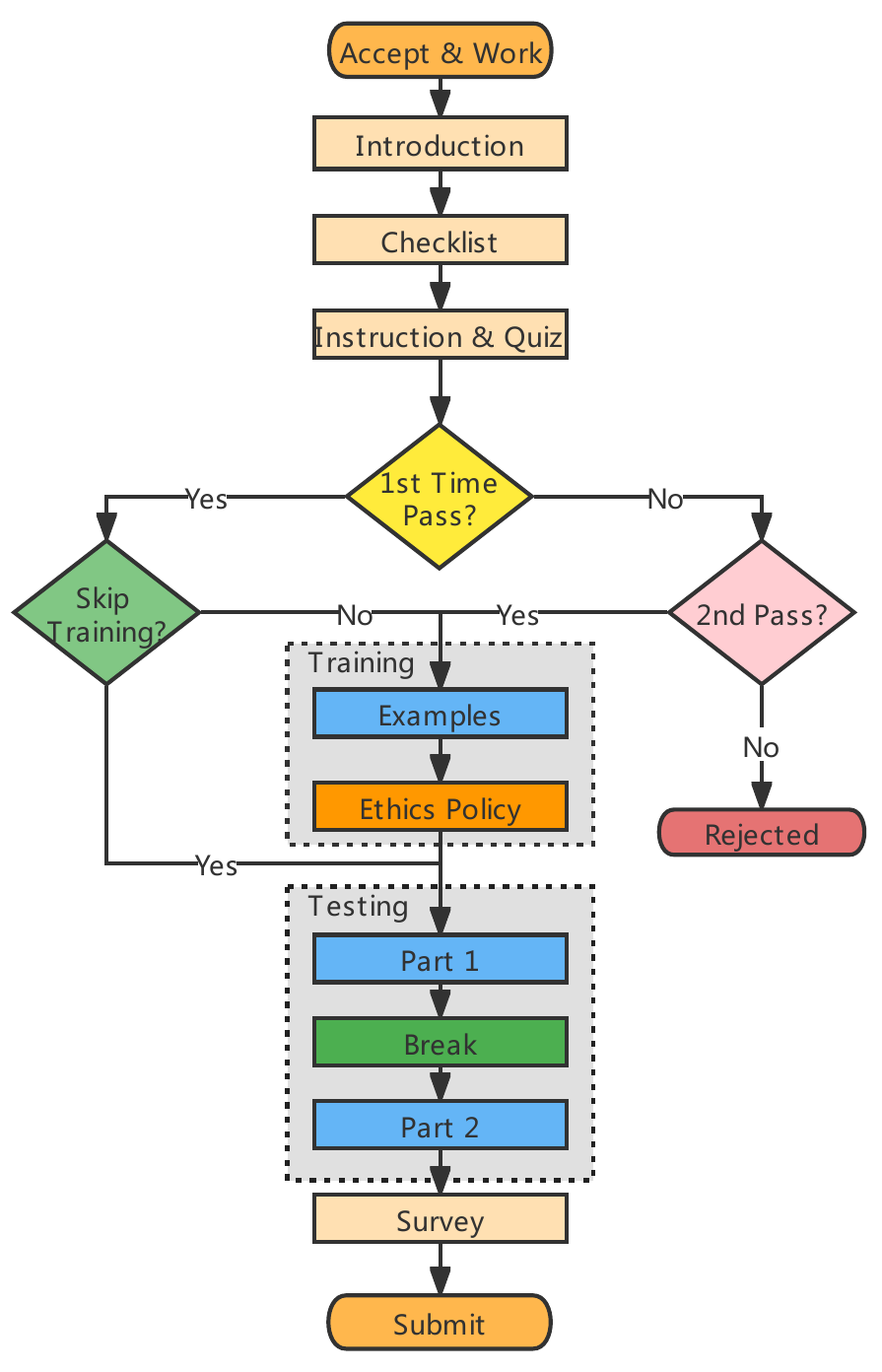}\hspace{1em}\includegraphics[width=0.14\textwidth]{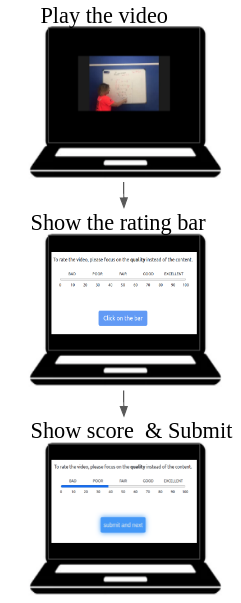}
\caption{\textbf{Left}: flowchart of the AMT workflow experienced by crowd-sourced workers when rating telepresence videos. \textbf{Right}: workflow when rating a video.}
\label{fig:flowchat}
\end{figure}

We used Amazon Mechanical Turk (AMT) to collect human opinions on the collected telepresence videos, as in earlier VQA studies ~\cite{livevqc,ytugc,paq2piq,clive,koniq}.
The human intelligence task (HIT) pipeline we designed is shown in Fig. \ref{fig:flowchat}. 
%
Each video was played without scaling on a black background, followed by a rating
interface that prompted subjects to record scores on a rating bar. Instead of the discrete Absolute Category Rating (ACR) scale used in prior AVQA studies, we used a continuous rating scale to allow the subjects to record quality judgements with greater freedom and increased sensitivity~\cite{discreteACRVsContinuous}. 
We cached each next video while each worker was rating a current video, and ensured that each video was entirely downloaded before playback to avoid rebuffering events and/or stalling. 

Next, we describe the overall study workflow in detail,
including protocols we employed to identify and eliminate unreliable subjects~\cite{livevqc,clive},
 or those having inadequate processing or network resources.

\noindent\textbf{Introduction:} 
We showed each subject a brief task description followed by 5 sample telepresence videos 
exemplifying a wide range of the possible quality
levels they might encounter in the study.
We only accepted workers with AMT acceptance rates $>$ 75\%. If a participant's browser window resolution, version, zoom, and the time taken to load videos did not meet our requirements, they were not allowed to proceed.

\noindent\textbf{Checklist:} We guided each subject to finish a series of setup steps to eliminate distractions, ensure their audio devices were on, adjust their seating, and wear their corrective lenses if necessary.

\noindent\textbf{Instruction and quiz:} To ensure the subjects' audio devices were on and working, we displayed the instructions,
then asked the subjects to 
reply to additional questions that were posed via audio.
The subjects were allowed to participants in several tasks, each time encountering a different set of videos. 

\noindent\textbf{Training:} A short training session was played that included 3 telepresence videos, to allow the participants to become familiar with the interface.
After that, each subject was required to display and read the ethics policy page for at least 30
seconds before proceeding. 
Although we ensured that each video was entirely downloaded prior to viewing, we also checked for any potential device-related video stalls. If the delay on any training video exceeded $2$ sec., or the total delay over the 3 training videos exceeded $3$ sec., the subject was not allowed to proceed (without prejudice). They were also stopped if a negative delay was detected (\eg, if they used plugins to speed up the video).

\noindent\textbf{Testing:} Each subject was asked to rate 90 videos. We divided each subject's participation into two sessions with a break between, to avoid fatigue. We analyzed the ratings
collected halfway through each session, and also at the end of the session, to identify unreliable workers.
At the middle of each subject's task, we checked for instability of the internet connection.
If more than $50\%$ of the videos viewed until then had suffered from hardware stalls, the subject was disqualified, without prejudice (they could try again). We also determined whether a subject had been giving very similar quality scores to all videos.


\noindent\textbf{Survey:} At the end of a subject's task, several survey questions were asked to collect subject demographics and to record video and audio device specifications. 



\subsection{Crowdsourcing Quality Control}

When conducting online crowd-sourced studies, quality control is essential to obtaining reliable quality labels. One common practice is to use repeated and/or ``golden'' videos for which the
highly reliable subjective scores were previously obtained, which may then be used  to compare with the worker's inputs. 

\noindent\textbf{Golden videos:} In prior VQA studies, ``Gold'' videos are often selected from other databases featuring similar content,
to provide reliable scores that can be compared against to detect dishonest workers.
Since there are no such existing databases containing reliable audiovisual telepresence contents and labels, 
we conducted a pilot study where we collected ratings on 90 videos, 
from among which we randomly selected 5 videos having a wide range of quality ratings. 
Those videos were inserted randomly into the first half-session videos of of each AMT task,
and were used to identify and eliminate unreliable subjects. 

\begin{figure}[t]
\centering
\includegraphics[width=0.3\textwidth]{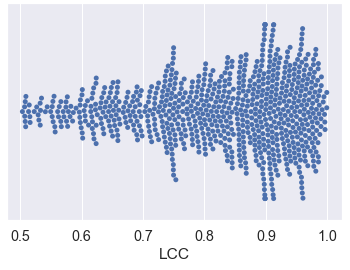}\includegraphics[width=0.32\textwidth]{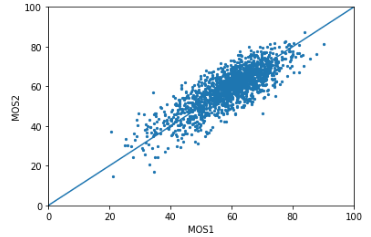}
\caption{\textbf{Left}: Swarm plot of individual LCC against ``golden videos'' illustrating high intrasubject consistency. \textbf{Right}: Scatter plot of a random 50\% division of the human labels into two disjoint subject sets illustrating high intersubject consistency.}
\label{fig:data_consistency}
\end{figure}

\noindent\textbf{Repeated videos:}
After half of the ratings were collected in each session, we systematically sampled 5 videos having diverse ratings, then inserted these randomly into the second half of the task. Based on the level of correlation of the ratings provided in the two half-sessions, we identified unreliable subjects and removed them from the study. 





\subsection{Subjective Recovery Analysis}
\label{sec:data_analysis}


Rather than applying the subjection rejection protocol BT.500 recommended by the ITU for video studies, 
which incorporates a number of hard coded parameters and thresholds, which may not be suitable for telepresence contents~\cite{li2020simple},
we instead adopted a recent ``soft'' subject rejection model~\cite{li2017recover} that is designed to recover  
subjective quality scores from noisy measurements.
To establish the internal integrity of the final set of collected subjective scores, we conducted two consistency checks on the recorded MOS, as shown in Fig.\ref{fig:data_consistency} and described in the following.


\noindent \textbf{Inter-subject consistency:} 
We randomly divided the subjects into two equal and disjoint sets and computed the
{Spearman Rank Correlation Coefficient (\textbf{SRCC})}
\cite{kendall1948rank} between the two sets of MOS over $50$ such random splits. We arrived at an average SRCC of \textbf{0.765}, 
indicating a high degree of agreement between the human subjects, implying successful testing and screening processes. A scatter plot of one of the divisions is shown in Fig.~\ref{fig:data_consistency} (right).

\noindent \textbf{Intra-subject consistency:} 
We also computed the 
{Linear Correlation Coefficient (\textbf{LCC})
\cite{pcc}} between the collected MOS, against the original scores on the ``golden" videos, obtaining a median LCC of \textbf{0.845}. 
These high correlations further validate the efficacy of our data collection process.





\section{TVQA Modeling}



TVQA algorithms should include two main modalities. As shown in Fig.~\ref{fig:system}, a quality measurement module in the application layer takes in rendered multimedia sources, providing feedback to control inputs that optimize the transport layer. 
This implies three requirements on the design of successful TVQA models. 
First, both video and audio quality need to be accurately modeled. 
While subjective experiments have shown that 
the visual component generally dominates overall audio-visual quality perception~\cite{common_avq_degradations}, 
audio quality is hardly insignificant (and in telepresence, is even more important),  and certain types of audio distortions (such as background noise and clipping) can cause the audio component to significantly impact the overall perception of quality. 
Second, a TVQA algorithm should be able to handle the ``missing-modality'' problem, whereby if either the video or audio signal
is not present during a video call, then the quality of the remaining signal is still accurately predicted. 
Third, a TVQA algorithm should be able to provide separate quality measurements on each modality as well as overall quality predictions. These can be useful for adjusting network traffic priorities for each modality~\cite{zoomDSCP}.
Based on these considerations, we first designed an image model called TeleIQA to effectively predict image quality both globally and locally. Then we integrated TeleIQA to a video model called TeleVQA to perform telepresence video quality assessment.

%


\subsection{Tele-IQA: our image model} 
We aim to build an image model that can efficiently give accurate predictions both on the full image and on the patches. This will enable modeling the spatial non-uniformity of telepresence content and providing predictions both locally and globally.  
IQA models can be viewed as mapping functions from the image domain to the real set. Our Tele-IQA model is a composition of three functions: feature extraction \textbf{f}, pooling \textbf{p}, and regression $\textbf{r}$.
%
To efficiently extract patch quality predictions, we use RoIPool to estimate local predictions on extracted feature maps instead of feeding patches to the network. The RoIPool operator ($\textbf{p}_{RoI}$) is designed to pool local feature maps from global feature maps:
\begin{equation}
    (\textbf{f}^{} \circ \textbf{p}_{RoI}) (P) \approx (\textbf{f}^{} \circ \textbf{p}) (P_{RoI}),
\end{equation}
where $P$ is a picture and $P_{RoI}$ is the local patch of $P$ correspoding to the region of interest (RoI). 
Therefore, the predicted patch quality score $\widetilde{S}_{RoI}$ is close to the prediction $S_{RoI}$ when taking the patch as input:
\begin{equation}
    \widetilde{S}_{RoI} = (\textbf{f}^{} \circ \textbf{p}_{RoI} \circ \textbf{r} ) (P) \approx (\textbf{f}^{} \circ \textbf{p} \circ \textbf{r}) (P_{RoI}) = S_{RoI}
\end{equation}

\begin{figure*}[tbh]
\centering
\includegraphics[width=0.7\textwidth]{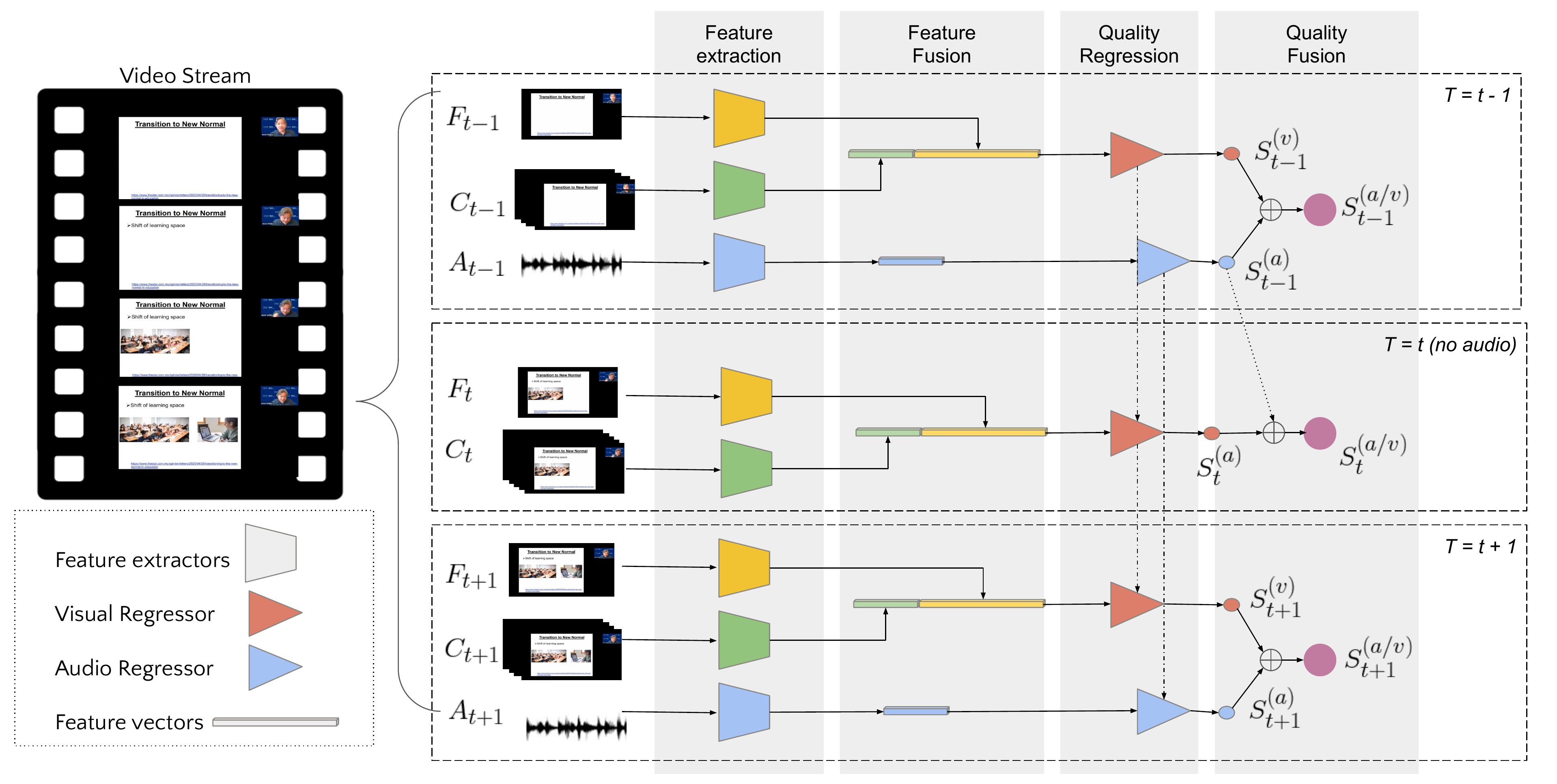}
\caption{
Our Tele-VQA model which involves 4 sequential steps:
feature extraction, feature fusion, quality regression, and quality fusion.
\mynew{For video conferences, the audio signal is not guaranteed to be always available. Here we describe how we handle the case of missing audio ($T = t$).}
}
\label{fig:teleVQA}
\end{figure*}

\subsection{Tele-VQA: our video model}

At each time step, Tele-VQA receives one frame ($F_t$), one video clip ($C_t$), and one audio clip ($A_t$), and generates timely visual ($S^{(v)}_t$), audio ($S^{(a)}_t$) and combined audio-visual quality predictions
 ($S^{(a/v)}_t,$).
Tele-VQA involves four sequential steps: feature extraction, feature fusion, quality regression, and quality fusion (\textit{c.f.} Fig.~\ref{fig:teleVQA}). First, features are extracted from patches, frames, video clips and audio clips from a video stream, capturing rich multi-modal information.
The patch, frame, and clip level features are fused and then fed to the visual regressor while the audio features are fed to the audio regressor.
Each regressor contain an internal state to allow information to flow from one time step to  
the next. Finally, the predicted visual and audio quality scores are fused to form an overall audio-visual quality prediction.
We provide more details of each step below.






\subsubsection{Feature Extraction} We use the frame backbone ($\textbf{f}^{(f)}$), clip backbone ($\textbf{f}^{(c)}$), and audio backbone ($\textbf{f}^{(a)}$) to extract features at the frame/patch, clip, and audio levels, respectively.

\begin{enumerate}
    \item \textbf{Frame-level features: $\mathcal{F}^{(f)}_t = (\textbf{f}^{(f)} \circ \textbf{p}) (F_t)$} For each incoming frame, 960 feature maps are computed using a MobileNetV3 backbone~\cite{howard2019mobilenetv3} pretrained on ImageNet~\cite{imageNet} and finetuned on the LIVE-FB Dataset~\cite{paq2piq}. Adaptive Average Pooling of size $1 \times 3$ is then applied, followed by flattening, yielding a 2880-dim feature vector.
    \item \textbf{Patch-level features:} Quality maps model spatial variations of distortions.  
To extract patch quality features, we partition the frames into a \mynew{$2^d\times 2^d$} grid of RoIs \mynew{($d = 1,2,3,4 $)} and apply RoI pooling on each RoI:
\begin{equation}
    \textbf{p}^{(M \times N)}_i(.) :=  \textbf{p}_{RoI_i}(.), \quad i \in \{1, 2, ..., M \times N \}.
\end{equation}
Then, the predicted quality scores for quality maps of different scales are concatenated into a vector: 
\begin{equation}
{\mathcal{F}}^{(p)}_t = \bigoplus^{4}_{d=1}   \bigoplus^{2^d \times 2^d}_{i=1} (\textbf{f}^{(f)} \circ \textbf{p}^{(2^d \times 2^d)}_i \circ \textbf{r} ) (F_t),
\end{equation}
where ${\mathcal{F}}^{(p)}_t$ is the extracted patch-level features on frame $F_t$;
$\bigoplus$ is the concatenation operator.
The extracted MobileNetV3 
features ($ 960 \times 3 \times 1$) are treated as a 960-variable time series of length of 3, which is then fed to a GRU-FCN~\cite{GRU_FCN}. 
Each extracted quality map is then flattened and concatenated into a 341-dim feature vector. 
    \item \textbf{Clip-level features: $\mathcal{F}^{(c)}_t = (\textbf{f}^{(c)} \circ \textbf{p}) (C_t)$} 
We employ a 3D CNN to extract spatiotemporal features to model video-level distortions such as flickering, jerkiness, and edge/texture floating~\cite{video_compression_artifacts}. 
We modified 
the R(2+1)D model~\cite{tran2018r2plus1d}
pre-trained on the Kinetics dataset~\cite{kinetics} by removing the last pooling layer, 
to serve as the backbone for extracting spatiotemporal features.
The backbone was not finetuned on quality-related tasks.
As in frame and patch level features, 
$1\times 3$ Adaptive Average Pooling is applied along the spatial dimension. Flattening then yields a 1536 dim feature vector.
    \item \textbf{Audio-level features: $\mathcal{F}^{(a)}_t = (\text{STFT} \circ \textbf{f}^{(a)} \circ \textbf{p} ) (A_t)$} Each  1D  audio  signal is  
transformed  into  a 2D 
spectrogram via  the  short-time  Fourier  transform  (STFT). 
Then, 2D features are extracted from each spectrogram using YAMNet~\cite{yamnet},
which is an audio classification model that incorporates a MobileNetV1
architecture that was pre-trained on the Google AudioSet dataset~\cite{googleAudioSet} to predict 521 different audio events. 
\end{enumerate}

\subsubsection{Feature Fusion}

We use separate pathways to process the visual and audio information. For the visual branch, we concatenate frame-level, patch-level, and clip-level features. The result is a 4757-dim feature vector for the visual pathway and a 1536-dim feature vector for the audio pathway.

\subsubsection{Quality Regression}

The resulting visual and audio features are fed to 
two different modified GRU-FCN~\cite{GRU_FCN} modules to conduct quality regression. 
The features extracted from the entire video may be viewed as a multi-variate time series.
Our online prediction model accepts a single sample point at each time step. 
{Quality regression is formulated as a Time Series Regression (TSR) problem, which we solve 
using GRU-FCN, a state-of-the-art deep model 
often used for Time Series Classification problem.}
GRU-FCN includes two main building blocks. 
The Gated Recurrent Unit (GRU) is used to learn temporal dependencies in a step-by-step manner,  while a fully convolutional network (FCN) accepts the entire time-series as input to perform feature extraction.
To adapt the GRU-FCN to the online quality regression problem, we set the input for the FCN to be the current sample point instead of the entire sequence. 


\subsubsection{Quality Fusion} 
We refer to ITU-T Rec. P.911~\cite{itu1999subjective} regarding how perceptions of audio and video quality ($S_t^{(a)}, S_t^{(v)}$) interact and how predictions of them can be combined into a single audio-visual quality prediction ($S_t^{(a+v)}$).
We add a fusion layer that uses the KPN model~\cite{belmudez2009KPN} to fuse the quality predictions: 
\begin{equation}
    S_t^{(a+v)} = S_t^{(a)} \otimes   S_t^{(v)} = 1.12 + 0.007 \cdot S_t^{(a)} + 0.24 \cdot S_t^{(v)} + 0.088 \cdot (S_t^{(a)} \cdot S_t^{(v)})
\end{equation}
%

{
We jointly trained the visual and audio pathways, and backpropagate the loss through the fusion layer, when both types of modality were available. 
When training Tele-VQA with a single type of modality, this fusion was not applied. 
To handle the ``missing modality'' problem during testing, Tele-VQA 
initializes both quality scores to $3.0$ (Fair quality) and then uses the last available quality predictions ($S^{(a)}_{t-1}$ or $S^{(v)}_{t-1})$,
as shown in Fig.~\ref{fig:teleVQA}. 
}

\subsubsection{Variants}  For the unimodal version of Tele-VQA, the extracted features for modality $m$ are fed into a TSR module $\textbf{r}_t^{(.)} \left[., .\right]$ to obtain a predicted score $S^{(m)}_t$:
\begin{equation}
    S^{(m)}_t = \textbf{r}_t^{(m)} \left[ \mathcal{F}_t^{(m)}  , \textbf{h}^{(m)}_{t-1} \right], \quad m \in \{ f, p, c, a \},\\
\end{equation}
where $\textbf{h}^{(m)}_{t}$ is the hidden state of the regressor $\textbf{r}$ for modality $m$ that captures historical information ($T = 1, 2, ..., t$).
The multimodal version of Tele-VQA for the video-only case is defined as:
\begin{align}
S^{(f+c)}_t &= \textbf{r}_t^{(fc)} \left[  \mathcal{F}^{(f)}_t\oplus \mathcal{F}^{(c)}_t , \textbf{h}^{(f+c)}_{t-1} \right]\\
S^{(p+f+c)}_t &= \textbf{r}_t^{(p+f+c)} \left[  \mathcal{F}^{(p)}_t\oplus \mathcal{F}^{(f)}_t\oplus \mathcal{F}^{(c)}_t , \textbf{h}^{(p+f+c)}_{t-1} \right].
\end{align}

Our final model considering all modalities is defined as:
\begin{equation}
S_t^{(p+f+c+a)} = S^{(p+f+c)}_t \otimes  S_t^{(v)}.
\end{equation}


\section{Experiments}


\mynew{We followed the common practice of leaving out 20\% for testing, while the remaining 80\% contains the training and validation data (train:val:test = 6:2:2).}
Unlike the way most deep image networks are trained, we did not crop, resize, rotate, or otherwise process the input videos. Any such operation would introduce additional spatial and/or temporal artifacts, making comparisons to human judgments of the non-altered video quality less meaningful. Processing input videos of diverse aspect ratios, resolutions, and durations, however, makes training an end-to-end deep network impractical. 
Therefore, we formatted the videos which are of highly diverse sizes for efficient training, using two regularization steps which we describe next. 

\noindent\textbf{Spatial regularization:} First,  extract video frame/clip features and audio mel spectrograms are computed and saved in a lossless compression format. The spatial dimensions of the audio spectrograms are all the same.
Before training, we converted all of the frames and clips of each video to a sequence of feature vectors.
In this way, we reduced the training time and also regularized the various spatial dimensions. 
Since frames and clips share the same spatial dimensions on each video, we fed them in batches into MobileNetV3 
and R(2+1)D, respectively.
We extracted features from all frames and all non-overlapping clips of 8 continuous frames in each video.

\noindent\textbf{Temporal regularization:} We used a fixed number of time steps for each video/audio so that we could feed them in batches to the regressors. 
To trade off feature coverage with training efficiency, we set the number of time steps to be 20 for the video and 10 for the audio. 
If fixed sampling of the videos/frames were used, we could not use all of the extracted features. 
Therefore, we instead used systematic random sampling to obtain different groups of evenly-spaced videos/frames during training. 
We introduced this randomization as a method of data augmentation to help avoiding over-fitting, by mapping different variants of videos to the same labels. 

\begin{table}[b!t]
\centering
\caption{\textbf{Parameter efficiency on the IQA task}: Performance when all models are trained and tested on the LIVE-FB dataset~\cite{paq2piq}. NIQE is not a trained model.
} 
\scriptsize
\begin{tabular}{l
|c|c||c|c
}
\hline
\textbf{Model}                   & \textbf{Year} & \textbf{\# Params} & \textbf{SRCC}  & \textbf{LCC}   \\  \hline
\textbf{Traditional models} & {} & {} & {} & {}  \\
~~~~NIQE~\cite{niqe}          &   2013   & -         & 0.094 & 0.131 \\ 
~~~~BRISQUE~\cite{brisque}        & 2012   & -         & 0.303 & 0.341 \\  \hline
\textbf{Deep models} & {} & {} & {} & {}  \\
~~~~CNNIQA~\cite{cnnIqa}         &  2014  & 0.72 M   & 0.259 & 0.242 \\ 
~~~~NIMA~\cite{nima}                 & 2018 & 2.23 M    & 0.521 & 0.609 \\ 
~~~~DB CNN\cite{zhang2018DBCNN}   & 2018 & 1.38 B    & 0.554 & 0.652 \\ 
~~~~P2P-FM~\cite{paq2piq}                  & 2020 & 12.24 M   & 0.562 & 0.649 \\ 
~~~~HyperIQA~\cite{su2020hyperIQA} & 2020 & 27.38 M   & 0.535 & 0.623 \\ 
~~~~MUSIQ~\cite{ke2021musiq}       & 2021 & 27 M*     & 0.566 & 0.661 \\ 
~~~~CONTRIQUE~\cite{pavanContrastiveLearning}               & 2021 & 27.97 M   & 0.580 & 0.641 \\ 
~~~~Tele-IQA (ours)          & -    & 9.79 M    & \textbf{0.580} & \textbf{0.675} \\  \hline
\end{tabular}
\label{tbl:parameter-efficiency}
\end{table}

\noindent\textbf{Implementation Details: } We built our code on the machine learning libraries Fast.ai2~\cite{howard2020fastai}
and tsai~\cite{tsai}.
We used a batch size of $128$ and employed the MSE loss when regressing the output quality scores.
We trained for 60 epochs with the Adam optimizer ($\beta_1=.9$ and $\beta_2=.99$) and a weight decay of $.01$, 
When finetuning MobileNetV3, we first froze the backbone, and tuned the head layers over $10$ epochs,
then we unfroze the backbone and followed a discriminative learning approach~\cite{howard2018universal} for 1 additional epoch, using a lower learning rate of $3e^{-4}$, but a higher learning rate of $3e^{-3}$ for the head layers. 







\subsection{Results}





\noindent\textbf{Image quality:} 
We evaluated Tele-IQA and leading IQA models, all trained on the 
existing LIVE-FB dataset~\cite{paq2piq},
and report their parameter efficiency (Table~\ref{tbl:parameter-efficiency}).
To observe whether trained models transfer well to other IQA databases,
Table~\ref{tbl:onFlive} reports cross-database validation results on two smaller, independent ``in-the-wild” databases, CLIVE~\cite{clive} and KonIQ~\cite{koniq} \underline{without any fine-tuning}. 

\begin{table*}[tbh]
\caption{\footnotesize{\textbf{Picture quality predictions: }
Performance of picture quality models on different databases~\cite{paq2piq}. A higher value indicates superior performance. 
}}
\centering
\scriptsize
\begin{tabular}
{l
||c|c
||c|c
}
\hline
& \multicolumn{2}{c||}{\textbf{CLIVE~\cite{clive} }} 
& \multicolumn{2}{c}{\textbf{KonIQ~\cite{koniq} }} 
\\
\hline
\textbf{Model} 
  & \textbf{SRCC} & \textbf{LCC} 
  & \textbf{SRCC} & \textbf{LCC} 
\\
\hline
NIQE \cite{niqe}        & 0.052 & 0.154 & 0.534 & 0.509 \\
BRISQUE \cite{brisque}  & 0.495 & 0.494 & 0.641 & 0.596\\
\hline
CNNIQA \cite{cnnIqa}    & 0.580 & 0.481 & 0.596 & 0.403\\
NIMA \cite{nima}        & 0.395 & 0.411 & 0.666 & 0.721 \\
\hline
P2P-FM ~\cite{paq2piq}  & 0.756 & 0.783 & \textbf{0.788} & \textbf{0.808} \\
\textbf{Tele-IQA}    & \textbf{0.767} & \textbf{0.795} & {0.772}	& {0.800}\\
\hline
\end{tabular}
\label{tbl:onFlive}
\end{table*}

\begin{table}[ht!]
\caption{\textbf{Video quality predictions}: Performance when all models are separately trained and tested on our database and LIVE-VQC. 
{Here p, f, c, a means patch, frame, clip, and audio features, respectively. }
} 
\centering
\scriptsize
\begin{tabular}
{l||c|c||c|c}
\hline
& \multicolumn{2}{c|}{\textbf{Our database} } & \multicolumn{2}{c}{\textbf{LIVE-VQC} \cite{livevqc}} \\

\hline
& \textbf{SRCC} & \textbf{LCC} & \textbf{SRCC} & \textbf{LCC} \\
\hline
\textbf{IQA models} & {} & {} & {} & {}  \\
{~~~~BRISQUE} \cite{brisque} & { 0.411 } & {0.482 } & 0.592 & 0.638 \\
~~~~TeleVQA (p)  & {0.476 } & {0.488} & {0.621 }	& {0.603 }
\\
~~~~TeleVQA (f)  & {0.609 } & {0.590 } & {0.710 }	& {0.716} 
\\	
	
\hline
\textbf{VQA models} & {} & {} & {} & {}  \\
~~~~VSFA \cite{vsfa} & {0.601} & {0.655} & 0.773 & 0.795 \\
~~~~TLVQM \cite{tlvqm} & {0.565} & {0.617} & 0.799 & {0.803} \\
~~~~VIDEVAL \cite{videval} & {0.536} & {0.560} & 0.752 & 0.751 \\
~~~~TeleVQA (c)  & {0.475} & {0.467} & {0.792 }	& {0.730} \\
~~~~TeleVQA (f+c)  & {0.621 } & {0.652 } & {0.811 }	& 0.801 \\
~~~~TeleVQA (p+f+c)  & {0.633 } & {0.672 } & \textbf{0.811 }	& \textbf{0.829} \\
\hline 
\textbf{AVQA models} & {} & {} & {} & {}  \\
~~~~TeleVQA (a)  & {0.114 } & {0.136 } & {- }	& {-} \\
~~~~TeleVQA (f+a)  & {0.622 } & {0.686 } & {- }	& {-} \\
~~~~TeleVQA (f+c+a)  & {0.639 } & {0.686 } & {- }	& {-} \\
~~~~TeleVQA (p+f+c+a)  & \textbf{0.663 } & \textbf{0.715} & {-}	& {-} \\
\hline 
	
\end{tabular}

\label{tbl:comp_vqa}
\end{table}


\noindent\textbf{Video quality:} 
Audio information is optional for Tele-VQA, making it possible to evaluate it
on the video-only LIVE-VQC database~\cite{livevqc}.
As shown in Table~\ref{tbl:comp_vqa}, 
Tele-VQA was able to compete very well with the other models on LIVE-VQC.


\noindent\textbf{Audiovisual quality:} 
Table~\ref{tbl:comp_vqa} compares the models' performances on our telepresence video database. 
%
Tele-VQA improved the SRCC 
by ${9.8\%}$ as compared to the best baseline. 
Adding the audio information contributes an improvement of ${2.4\%}$.
We also studied the contribution of the features from different modalities on the performance, 
by training separate models with one or two modalities  excluded.  
As may be observed from this ablation, 
integrating multimodal features contributed to a significant performance boost on both telepresence videos and UGC videos. 
Comparing unimodel Tele-VQAs,
image-level features contributed more than patch-level features.
Clip-level features were essential for modeling UGC videos 
but contributed the least to the telepresence. 

\begin{figure}[htb!]
\centering
\includegraphics[width=0.51\textwidth]{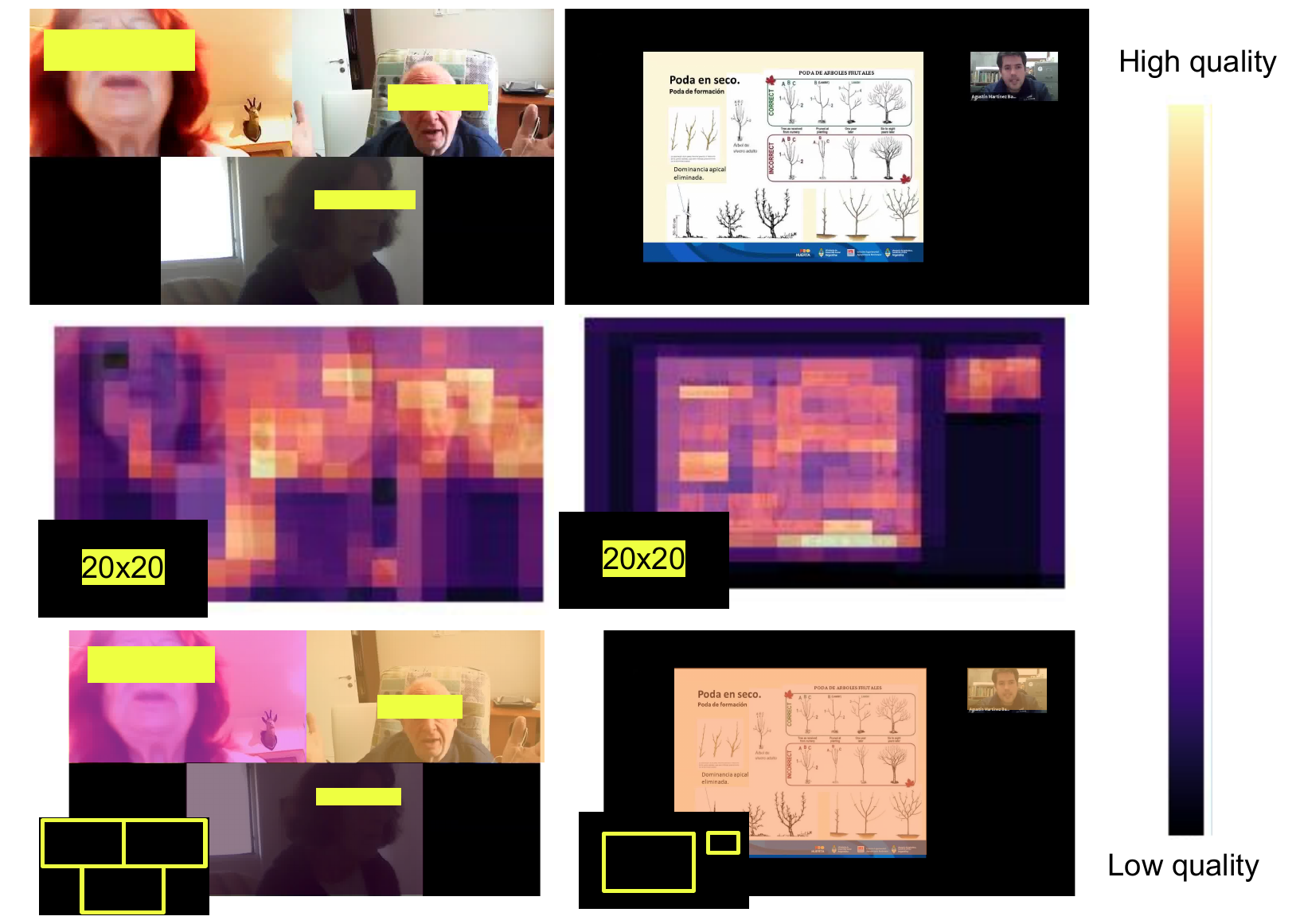}
\caption{Example of applying Tele-VQA on telepresence frames (top) to extract $20\times 20$ quality maps (middle) and 
local quality predictions on selected regions (bottom).}
\label{fig:application_local_quality}
\end{figure}

\noindent\textbf{Provide local quality predictions:} 
Telepresence videos often exhibit spatial non-uniformities of visual quality, \eg when in grid view or slide-sharing view, where the videos are from different capture, compression and transmission sources. 
\mynew{To monitor the visual quality of interested regions, we feed their coordinates of the in addition to the multi-scale grids of RoIs when extracting patch-level
features.}
As shown in Fig.~\ref{fig:application_local_quality}, Tele-VQA is able to provide  multi-scale quality maps, as well as perceptual quality predictions on regions of interest. Integrating Tele-VQA into a video conferencing platform in this way can provide timely local quality feedback, 
which can be used to support the development of optimal strategies for processing different sources of videos and pictures.


\section{ Discussion and Conclusion}


Videoconferencing has become much more popular in
recent years, so much so that it has altered the way many people communicate. Monitoring the perceptual quality of a video conferencing session having multiple participants using different types of equipment can present many complexities. 
To ensure
the best possible quality, standardized methods to quantify  audiovisual teleconferencing quality
of telemeetings are needed. 
Although  many other studies  have  addressed  video and  audio  quality,  
few have  simultaneously  addressed both, and none have in the context of video teleconferencing.
This is unfortunate, since both types of sensory signals shape 
user-perceptions of the quality of teleconferencing sessions.
To help address these challenges, we built a new dedicated telepresence quality dataset that is substantially larger, more diverse, and more representative of video conference signals than any previous audiovisual datasets.
We also created an all-in-one audio-visual quality prediction model, called Tele-VQA,
which integrates multimodal features to accurately infer 
telepresence image,video, audio and audiovisual quality. 
It is also able to generate predictions in an online manner,
while achieving  has shown state-of-the-art performance both on our new database and on other datasets.

While Tele-VQA was designed for
in telepresence applications, 
further applications and extensions are possible. 
One avenue for future work is 
to assist visually challenged users 
to take better quality videos by 
providing timely video quality feedback as guidance. 
Another very interesting line of inquiry would be to consider the impact on audio-visual quality of space-time saliency (\eg a current speaker voice and image) 
towards designing and 
integrating visual attention mechanisms to further improve the prediction accuracy of Tele-VQA. 



\subsubsection{Acknowledgments:} This work was supported by Meta Platforms, Inc. A.C. Bovik was supported in part by the National Science Foundation AI Institute for Foundations of Machine Learning (IFML) under Grant 2019844.

\clearpage
%
%
\bibliographystyle{splncs04}
\bibliography{bib_vqa}

\end{document}